\documentclass[10pt,twocolumn,letterpaper]{article}

\usepackage{authblk}

\usepackage[pagenumbers]{cvpr}
\usepackage{multirow}
\usepackage{array}
\usepackage{bbm}
\usepackage{graphicx}
\usepackage{amsmath}
\usepackage{amssymb}
\usepackage{booktabs}
\usepackage{bbding}
\usepackage{color}
\usepackage{appendix}
\usepackage{fancyhdr}
\usepackage{CJKutf8}

\usepackage[dvipsnames]{xcolor}

\definecolor{cvprblue}{rgb}{0.21,0.49,0.74}
\usepackage[pagebackref,breaklinks,colorlinks,citecolor=cvprblue]{hyperref}
\usepackage{xcolor}
\definecolor{DarkGreen}{rgb}{0.43, 0.68, 0.28}

\newcommand\keywords[1]{\textbf{Keywords}: #1}

\title{RemoCap: Disentangled Representation Learning for Motion Capture}

\author[1,2]{Hongsheng Wang\textsuperscript{*}}
\author[2]{Lizao Zhang\textsuperscript{*}}
\author[2]{Zhangnan Zhong}
\author[2]{Shuolin Xu}
\author[2]{Xinrui Zhou}
\author[1]{Shengyu Zhang} 
\author[3]{Huahao Xu \textsuperscript{$\dag$}}
\author[1]{Fei Wu}
\author[2]{Feng Lin}

\affil[1]{Zhejiang University, China}
\affil[2]{Zhejiang Lab, China}
\affil[3]{Gameday Inc., China}

\makeatletter
\renewcommand\AB@affilsepx{, \protect\Affilfont}
\makeatother

\begin{document}

\maketitle

% \thispagestyle{fancy}

% \lfoot{This work has been submitted to the IEEE for possible publication. Copyright may be transferred without notice, after which this version may no longer be accessible.}

% \cfoot{}

\renewcommand{\headrulewidth}{0mm}

\begin{CJK}{UTF8}{gbsn}
\renewcommand{\thefootnote}{\fnsymbol{footnote}}
\footnotetext[1]{These authors contributed equally to this work.}
\footnotetext[2]{Corresponding Author.}
\begin{abstract}

Reconstructing 3D human bodies from realistic motion sequences remains a challenge due to pervasive and complex occlusions. Current methods struggle to capture the dynamics of occluded body parts, leading to model penetration and distorted motion. \textbf{RemoCap} leverages \textbf{S}patial \textbf{D}isentanglement (\textbf{SD}) and \textbf{M}otion \textbf{D}isentanglement (\textbf{MD}) to overcome these limitations. \textbf{SD} addresses occlusion interference between the target human body and surrounding objects. It achieves this by disentangling target features along the dimension axis. By aligning features based on their spatial positions in each dimension, \textbf{SD} isolates the target object's response within a global window, enabling accurate capture despite occlusions. The \textbf{MD} module employs a channel-wise temporal shuffling strategy to simulate diverse scene dynamics. This process effectively disentangles motion features, allowing RemoCap to reconstruct occluded parts with greater fidelity. Furthermore, this paper introduces a sequence velocity loss that promotes temporal coherence. This loss constrains inter-frame velocity errors, ensuring the predicted motion exhibits realistic consistency. Extensive comparisons with state-of-the-art (SOTA) methods on benchmark datasets demonstrate RemoCap's superior performance in 3D human body reconstruction. On the 3DPW dataset, RemoCap surpasses all competitors, achieving the best results in MPVPE (81.9), MPJPE (72.7), and PA-MPJPE (44.1) metrics. Codes are available at \url{https://wanghongsheng01.github.io/RemoCap/}.

\end{abstract}


\section*{Introduction}\label{sec:intro}
\
The reconstruction of precise and temporally stable 3D human mesh sequences from video has broad applications across various industries, including medical rehabilitation, game development, sports motion analysis, and clothing design~\cite{tian2023hmrsurvey}. However, this task presents a significant challenge when dealing with complex occlusion scenarios.

Current methods for 3D human reconstruction can be categorized into two main approaches: spatial and temporal. Spatial methods leverage pixel-aligned local cues like 3D keypoints~\cite{iqbal2021kama,li2021hybrik}, mesh vertices~\cite{moon2020i2l}, and mesh-aligned features~\cite{zhang2021pymaf}. While these methods achieve accurate localization, they struggle with occlusions as the underlying image information becomes unavailable. In contrast, temporal methods, including~\cite{shen2023global}, incorporate motion information to improve intra-frame accuracy. However, due to the high degree of interdependence between features, these methods struggle to capture motion in severely occluded regions. This limitation often leads to model penetration artifacts in the reconstructed body parts.

\begin{figure*}[h]
  \centering
    \includegraphics[width=0.9\linewidth]{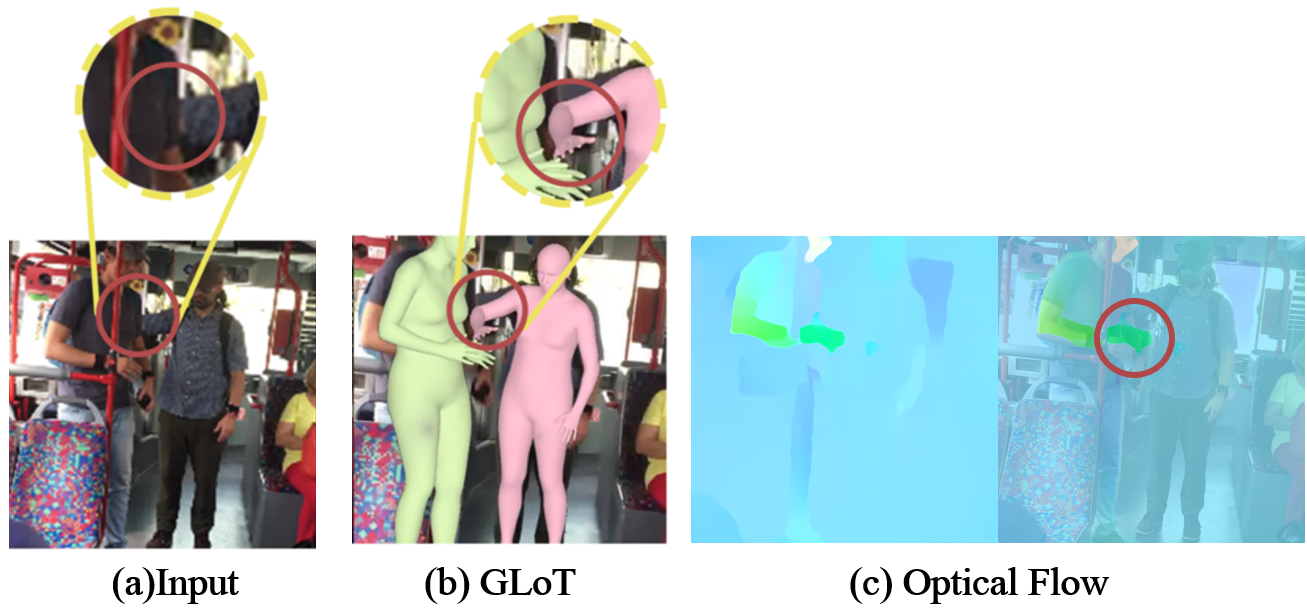}  
\caption{(a) Image input with occlusion, (b) Distortion in the reconstruction of occluded parts by GLoT~\cite{shen2023global}, (c) Optical flow information for the scene.}
\label{fig:i1}

\end{figure*}

As illustrated in Figure ~\ref{fig:i1}, feature coupling arises during occlusion. This occurs when interactions between the target object (human) and non-target objects (\textit{e.g.}, another person) lead to confusion about the target's features.  For example, occlusion between two people might cause the model to mistakenly use the left person's arm to reconstruct the right person's arm. To bridge the gap, we introduce the \textbf{S}patial \textbf{D}isentanglement (SD) module. This module effectively decouples the target's features, enabling the learning of spatially consistent features specific to the target object. By mitigating the interference of non-target features caused by occlusion, RemoCap significantly improves the reconstruction process in such scenarios.

Occlusion of target objects is often caused by motion in complex environments. When the target object moves, occlusions between body parts can occur. This leads to a phenomenon called "motion feature coupling," where local features become entangled. As illustrated in Figure ~\ref{fig:i1}, using methods like Flownet~\cite{dosovitskiy2015flownet} for optical flow extraction can suffer from this issue. In regions with dense motion, occlusions disrupt the target's true motion patterns, leading to distorted reconstruction results. To address this challenge of motion feature coupling during occlusions, we introduce the \textbf{M}otion \textbf{D}isentanglement (\textbf{MD}) module. This module specifically aims to capture the inherent consistency of the target object's motion across its body parts. By achieving this, the MD module effectively decouples the local motion features, enabling more accurate reconstruction.

Monocular image-based human mesh reconstruction falls into two main categories: model-based and model-free methods. Model-based approaches~\cite{loper2015smpl,li2023hybrik,zhang2021pymaf,tian2023hmrsurvey} rely on regressing a set of parameters (SMPL~\cite{loper2015smpl}) that represent human body pose and shape. These methods require few parameters and are widely applicable due to their ability to fit diverse image data. Conversely, model-free methods~\cite{lin2021end,lin2021mesh,cho2022cross} bypass pre-defined models and directly predict 3D mesh vertex coordinates from image features. They exploit the inherent correlations between human joints and mesh vertices across various poses, making them particularly well-suited for addressing the "feature coupling" issue presented in this work. Additionally, model-free methods often achieve superior single-frame reconstruction accuracy compared to their parametric counterparts. Therefore, this paper leverages the advantages of model-free methods to tackle occlusions effectively in the context of feature coupling.

RemoCap tackles the challenge of distorted motion reconstruction arising from coupled spatial and motion features, as well as the interference between target and non-target object features during occlusions. To ensure robust reconstruction of occluded motion, RemoCap leverages two novel modules: the Spatial Disentanglement Module (SD) and the Motion Disentanglement Module (MD). These modules effectively decouple spatial and temporal features, respectively. Extensive experiments on the 3DPW benchmark conclusively demonstrate RemoCap's superiority in 3D human mesh reconstruction. Our method outperforms state-of-the-art approaches on all evaluation metrics, establishing a new benchmark for this challenging task.

The primary contributions of this paper are summarized as follows:

\begin{itemize}
  \item We propose RemoCap, a novel Transformer-based framework for model-free human mesh reconstruction.
  \item We introduce two innovative modules: (i)The Spatial Disentanglement Module (SD) addresses the coupling of target and non-target features, enabling accurate reconstruction even in occluded scenarios. (ii)The Motion Disentanglement Module (MD) disentangles motion features, ensuring temporally consistent and accurate motion capture.
  \item We introduce a novel sequence velocity loss that aligns with temporal characteristics, further enhancing temporal consistency in the reconstructed sequence.
\end{itemize}

\section{Related Works}\label{sec:Related Work}

\subsection{Recovery From Monocular RGB Images}

Significant progress has been achieved in 3D human mesh reconstruction from monocular RGB images. Two main research directions have emerged: parametric model-based~\cite{bogo2016keep,  georgakis2020hierarchical, kolotouros2019learning, omran2018neural, kocabas2021pare} and model-free approaches~\cite{lin2021end,moon2020i2l}.

\textbf{Parametric Model-Based Approach:}This approach focuses on estimating pose and shape parameters of the SMPL model~\cite{loper2015smpl} to reconstruct the complete 3D human mesh vertices. Works like ~\cite{bogo2016keep, lassner2017unite} and the SMPL model itself ~\cite{loper2015smpl} are key examples. Additionally, Generative Adversarial Networks (GANs) have shown promise, with Kanazawa et al.'s work ~\cite{kanazawa2018end} and PyMAF ~\cite{zhang2021pymaf} being notable examples. However, while effective, these approaches are primarily suited for human body reconstruction and may face challenges with other models.

\textbf{Model-Free Approach:}This approach aims to directly reconstruct the complete vertices of a 3D human mesh, reducing reliance on the parameter space limitations of model-based methods. Pioneering works in this field include those by Kolotouros et al. ~\cite{kolotouros2019convolutional} and Choi et al. ~\cite{choi2020pose2mesh, ye2021sparse, velivckovic2017graph}. Recent advancements include the introduction of Transformer encoders by Lin et al. ~\cite{lin2021end} and the adoption of encoder-decoder architectures by Cho et al. ~\cite{cho2022cross} to disentangle the complex relationship between image features and grid queries.

\subsection{Recovery From Monocular Videos}

\textbf{Vertex-Based Approach:}This approach directly regresses the 3D vertex coordinates of the deformed mesh ~\cite{cho2022cross, choi2020pose2mesh,lin2021mesh}. GraphCMR ~\cite{kolotouros2019convolutional} leverages a graph convolutional network (GCN) constructed with an adjacency matrix to preserve the mesh's topological structure. The Transformer architecture has been successfully applied with methods like Mesh Transformer (METRO) ~\cite{lin2021end}.

\textbf{Temporal Modeling:}To better consider the temporal dimension of videos, most methods incorporate a temporal encoder to integrate features from each frame. HMMR ~\cite{kanazawa2019learning} employs a convolutional encoder for features extracted from HMR ~\cite{kanazawa2018end}. VIBE ~\cite{kocabas2020vibe}, MEVA ~\cite{luo20203d}, and TCMR ~\cite{choi2021beyond} leverage recurrent temporal encoders, while MAED ~\cite{wan2021encoder} and t-HMMR ~\cite{pavlakos2022human} adopt Transformer-based encoders. However, these methods often struggle in complex scenarios like heavy occlusions, leading to unsatisfactory results when feature information is missing on sequence segments ~\cite{kanazawa2019learning,zhang2019predicting}.

Subsequent works have addressed these limitations ~\cite{choi2021beyond, kocabas2020vibe, luo20203d, lee2021uncertainty, sun2019human, pavlakos2022human, vaswani2017attention, wan2021encoder, guan2021bilevel,wang2023selfannotated}. GLoT et al. ~\cite{shen2023global}. propose long and short-term modeling of time motion decoupling with Transformers. Issues like occlusion and jitter highlight the importance of optimizing methods for recovering 3D human body shapes from videos ~\cite{zeng2022smoothnet,zeng2022deciwatch,Zhao_2023_CVPR, rajasegaran2021tracking, rajasegaran2022tracking, cho2022cross}. These methods exploit human motion priors in the time series information to complete missing features and smooth keypoint positions, leading to optimized pose sequences, but potentially sacrificing some accuracy.  Additionally, their focus on keypoint coordinate completion often lacks mesh supervision, making the parametric model-based approach more suitable for video-based 3D human mesh recovery.

\begin{figure*}[ht]
  \centering
    \includegraphics[width=0.9\linewidth]{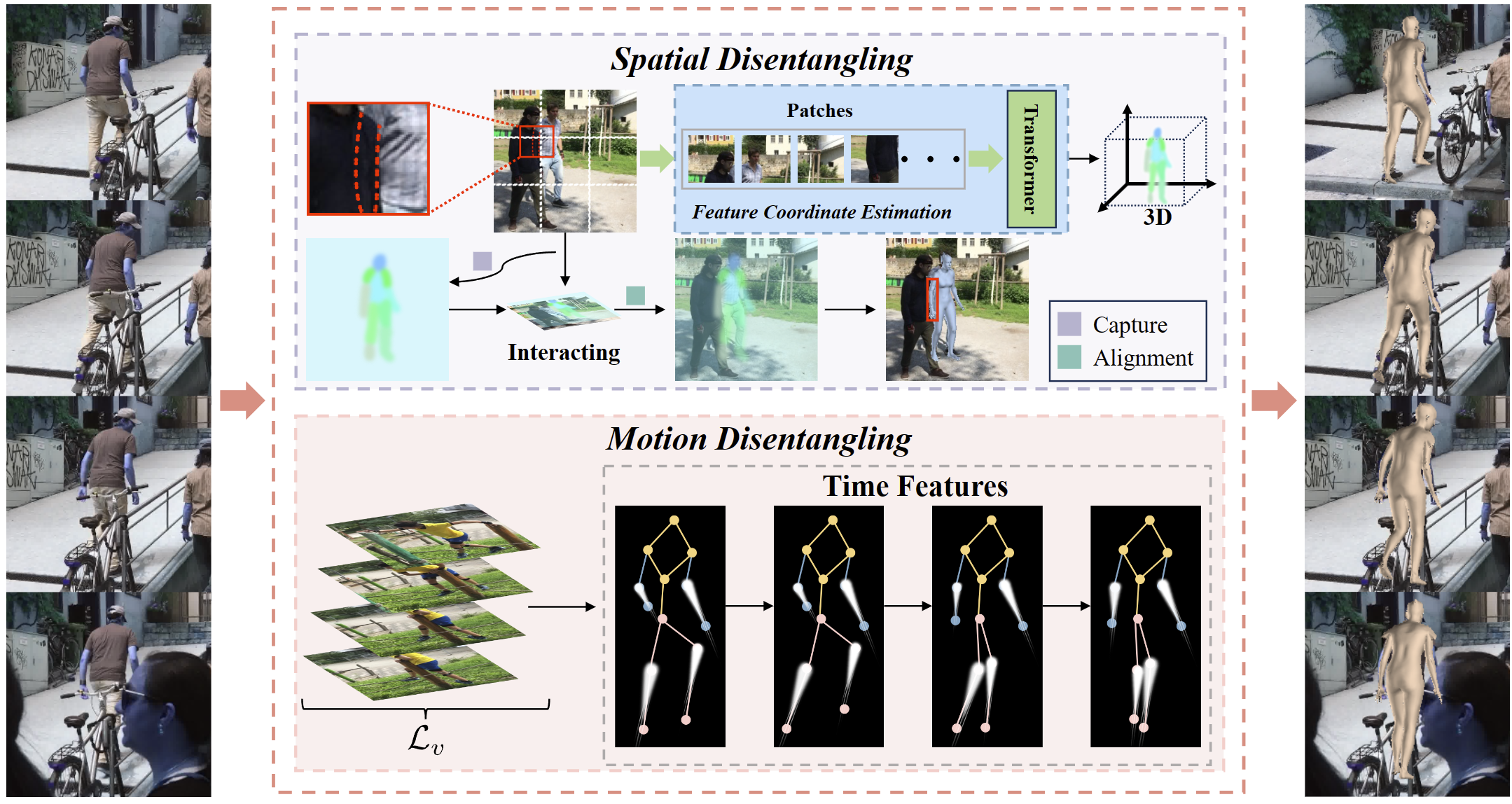} 
\caption{This figure illustrates the pipeline of the RemoCap model. Feature maps first undergo disentanglement by the Spatial Disentanglement (SD) and Temporal Disentanglement (TD) modules. The disentangled features are then reweighted using a sigmoid function before being decoded by a Transformer encoder to generate the final sequence of 3D human mesh vertices.}
\label{fig:pipeline}
\vspace{-4mm}
\end{figure*}

\section{Methodology}
\vspace{-1mm}
\label{sec:Methodology}
This section explores RemoCap's architecture, which consists of two crucial modules: the Spatial Disentanglement (SD) Module and the Motion Disentanglement (MD) Module (Figure ~\ref{fig:pipeline}). The SD Module tackles occlusions by precisely extracting features of the target object, even when obscured. It achieves this by leveraging a progressive transformer for accurate 3D coordinate estimation. The MD Module differs from the SD Module by focusing on extracting features across different spaces within the data. It prioritizes two key tasks: fusing temporal information from the video sequence and decoupling the target object's spatial features from its motion features. This decoupling process is essential for effectively addressing the challenge of "motion feature coupling" during occlusions.
\vspace{-2mm}

\subsection{Spatial Disentanglement}
Video reconstruction windows often contain features from both the target object (\textit{e.g.}, human body) and background elements. Accurately recovering 3D human mesh vertices requires learning consistent features that specifically correspond to the target object within this mixed feature space. To address this challenge, we propose a learning framework consisting of \textbf{spatial localization}, \textbf{spatial interaction}, and \textbf{spatial alignment}. By following these steps, our framework learns consistent features that accurately map to the target object's spatial location, enabling precise 3D mesh recovery.

\textbf{Spatial localization: }The reconstruction window often contains features from both the target object (\textit{e.g.}, human body) and background clutter. To isolate the target object's features (spatial target features), we employ a multi-dimensional attention mechanism channel by channel. This mechanism learns to focus on the target features within the spatial feature map, guided by the ground truth human body mesh vertex labels. By analyzing correlations across different feature dimensions, the attention mechanism amplifies the target features' contribution to the overall spatial representation. Conversely, the weights assigned to non-target features are adaptively minimized.

Specifically, given an input feature map$X \in \mathbb{R}^{C \times H \times W}$, this module partitions the input feature $X$ into $G$ groups along the channel dimension, each group containing $C/G$ channels. This division is designed to facilitate the learning of diverse feature representations, enhancing feature diversity. The sub-feature groups can be represented as  $X = [X_0, X_1, \ldots, X_{G-1}], X_i \in \mathbb{R}^{C/G \times H \times W}$, and the input channel count is transformed from $C$ to $C/G$. Following channel rearrangement, the module separates the features of the target object from the current reconstruction window, which is entangled with non-target object features along the $H$ direction. ($H$ represents the height of the feature map.) Technically, $W$ (the width of the feature map) is discretized into $H$ equal parts, retaining precise location information along the $H$ direction. The module learns attention weights for each equal part of $W$ corresponding to the position along the $H$ direction, reinforcing the response of target features in the global spatial features of the reconstruction window along the $H$ dimension. Global average pooling along the $H$ direction for global information in C/G can be expressed as:

\vspace{-1mm}
\begin{equation}
    z'_c(H) = \frac{1}{W} \sum_{i=0}^{W} x_c(H, i) 
\end{equation}

Here, $x_c$ represents the input feature of the $c$-th channel. Another pathway directly utilizes one-dimensional global average pooling along the $W$ direction (vertical dimension). This essentially generates a collection of positional information along this axis. This pathway then employs one-dimensional global average pooling along the $H$ direction (horizontal dimension). This step aims to capture spatial weights (importance) over long distances within the feature map while critically preserving precise location information along the $W$ direction. To further focus on the target object, the module learns attention weights for each equal part of $H$, corresponding to its position along the $W$ direction. This process enhances the response of target features within the global spatial features of the reconstruction window, specifically along the $W$ dimension. Finally, global average pooling is applied along the $W$ direction to capture global information. This can be mathematically expressed as:

\vspace{-1mm}
\begin{equation}
    z'_c(W) = \frac{1}{H} \sum_{i=0}^{H} x_c(W, i) 
\end{equation}
This section defines $x_c$ as the $c$-th channel of the input feature map. The module begins by encoding global information within this feature map. This encoding process strengthens the response of target object features within the reconstruction window's global spatial representation, along both the horizontal $H$ and vertical $W$ directions. By leveraging accurate spatial positions, this step effectively disentangles the target object's features from those belonging to the background.

\begin{figure}[h]
  \centering
    \includegraphics[width=1\linewidth]{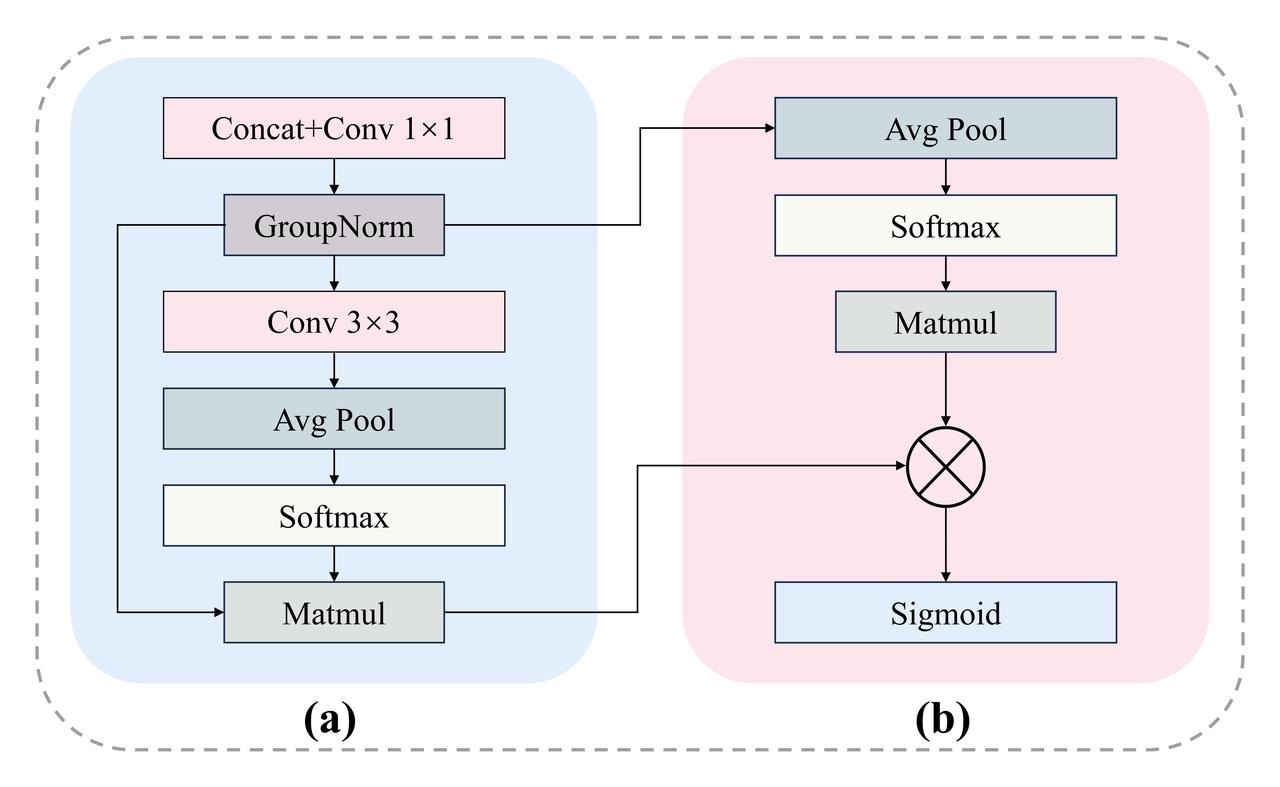} 
\caption{Feature Disentanglement Module Internal Details.}
\label{module}
\end{figure}

\textbf{Spatial interacting: }This section describes the network structure for spatial interaction, designed to analyze each channel of the feature map at different scales. This process strengthens the consistency of spatial features belonging to the target object, particularly along the channel dimension (refer to Figure ~\ref{module}(a)). Module (a) takes the concatenation of $z'_c(H)$ and $z'_c(W)$ as input. Here, $z'_c$ represents the processed feature map after some operation. A 1x1 convolutional kernel is used within this module to identify correlations between features across different channels. In parallel, another branch utilizes 3x3 convolutional kernels to capture local interactions between channels, ultimately improving the model's ability to express these features effectively. The final step involves fusing the channel-level attention maps generated by both modules (a) and (b). This fusion not only strengthens the encoding of information between channels but also critically preserves the integrity of precise spatial structural information within each feature channel.

\textbf{Spatial Alignment: }Having established correlations between target features across channels, we aim to further amplify their response throughout the entire feature space. To achieve this, we propose learning a spatial attention map. This map strengthens the signal from the target object's features at specific spatial locations within the reconstructed object. Essentially, it captures long-range dependencies and encodes global information. Module (a) constructs this spatial attention map by encoding global spatial information. It achieves this via 2D global average pooling applied to the output of the 1x1 convolutional branch (refer to Figure ~\ref{module}(a) for module structure). The output from another branch (minimal branch) is directly transformed to match the dimensional shape required for the subsequent channel-wise joint activation step. Finally, the original intermediate feature map is integrated into the final output.

This module offers two key advantages:(1)Preserves Spatial Information: It maintains accurate spatial position information throughout the process. (2)Learns Consistent Features: It efficiently learns consistent features for spatial alignment by considering both the consistency between features across channels and their corresponding spatial locations.

\subsection{Temporal Disentanglement}
The SD module excels at extracting spatial features of the target object from the reconstruction window. However, it has limitations in capturing long-range motion relationships between body parts. Recent advancements in cross-space and cross-channel analysis ~\cite{hou2021coordinate,10096516} motivated us to incorporate MD into our framework.MD addresses this challenge by treating consecutive video segments as a whole for analysis, using a technique called sequence shuffling. This shuffling process introduces diversity at the channel level, essentially simulating various temporal combinations. By analyzing the overall spatial features of these shuffled segments, MD extracts target motion features by examining changes along the temporal dimension. As a result, MD achieves a crucial step: decoupling motion features from static features across time, space, and channels.

Our focus now shifts to isolating the feature space of the target object within motion segments. To achieve this, we separate the motion features along the temporal dimension from the overall feature representation.
Specifically, for any given input feature map \(X \in \mathbb{R}^{(BS) \times C \times H \times W}\), where $B$ represents the batch size, $S$ denotes the number of frames grouped together and input to the network (in this case, $S$=8), and $H$ and $W$ respectively represent the spatial dimensions of the input feature. By restructuring the data into \(X \in \mathbb{R}^{(BC) \times S \times H \times W}\), we transform the format to highlight the temporal dimension. This essentially isolates each channel's feature maps across the video segment $S$. These temporally isolated features are denoted as \(X = [X_0, X_1, \ldots, X_{C-1}]\), where each \(X_i \in \mathbb{R}^{S \times H \times W}\) represents the feature map for a specific channel across $S$ frames.

Building upon the spatial alignment features learned in module (b) (refer to Figure ~\ref{module}) and the spatial attention maps extracted from the temporal features at various scales (1x1 and 3x3), this module achieves two key objectives:(1)Preserves Spatial Information: It incorporates the spatial attention weights into each group of output feature maps. This ensures that precise spatial position information is maintained throughout the process. (2)Harnesses Global Motion Information: By effectively combining the feature maps with their corresponding attention weights, the module efficiently utilizes global motion information within the video sequence.
The combined efforts of this module and the SD module lead to our approach's core strength: decoupling. We can effectively separate the spatial information of the target object from background clutter, while simultaneously isolating the target object's motion features from its static features. This leads to temporally consistent features, particularly beneficial in scenarios where features are coupled (\textit{e.g.}, occlusions). Overall, this approach demonstrates exceptional performance and practical value.

\textbf{Vertex Loss: }Previous research has shown that incorporating frame-to-frame velocity errors into the loss function improves the motion consistency of predicted results ~\cite{choi2020pose2mesh,DwivediDSR2021,kanazawa2018end}. Our proposed MD separates motion and static features within a video clip. To leverage this separation during training, we introduce a loss function that focuses on inter-frame motion features within the same video clip.
Specifically, we calculate the velocity changes of human body mesh vertices only between frames belonging to the same video segment. This ensures that during evaluation, only the temporal changes within a single sequence are considered. This focus on intra-sequence velocity facilitates the MD's ability to distinguish between temporal inter-frame motion features and static features.
The mathematical formulation for the sequence velocity loss function is provided below:

\begin{align}
\mathcal{L}_{v} &= \frac{1}{S} \sum_{s=1}^{S} \lVert V_{predict} - V_{gt} \rVert \\
V_{predict} &= \frac{1}{KT} \sum_{t=2}^{T} \sum_{i=1}^{K} \lVert J_{t,i} - J_{t-1,i} \rVert_2
\end{align}

where \( S \) represents different sequences, \( T \) represents the number of frames, \( J \) represents the number of key points, and \( K \) represents the number of key points.

This learning process leverages velocity errors between corresponding points in different frames of the same video. By analyzing these errors, the attention mechanism essentially learns to identify features that change over time (dynamic features). Conversely, the model assigns minimal weight to features that remain constant throughout the sequence (static features). Building on these temporally-aware features, we can further differentiate between the target object's features and background clutter within a single video sequence. This approach effectively discriminates between the target and non-target features by considering both the temporal information (changes across frames) and the spatial information (location within the frame). This enables us to precisely reconstruct the target object's features even in complex video sequences with intricate motions.

Our approach leverages the MD and supervises velocity errors within video segments. This strengthens the learning process for target features across various spatial dimensions and time sequences. To further differentiate between static and dynamic features within frames, we employed multi-scale interaction learning. This enhances the consistency of target features across time.  Furthermore, we obtain multi-scale attention maps, enabling effective feature fusion and leading to richer feature representations.

Our approach achieves significant improvements in target feature representation through the combined action of two modules. The first module separates the spatial information of the target object from the background clutter. This effectively enhances the focus on the target object's details.  The second module disentangles motion features from static features within the target object itself. This leads to temporally coherent features, particularly beneficial in complex video sequences with intricate motions. Overall, this decoupling strategy underpins our approach's exceptional performance and practical value in these challenging scenarios.

\subsection{Loss function}

Our model is trained using a combination of three loss functions: vertex loss (\(\mathcal{L}_{v}\)), 3D joint loss (\(\mathcal{L}_{3D}\)), and 2D joint loss (\(\mathcal{L}_{2D}\)). These losses are all calculated using the L1 distance metric. The overall training loss function is formulated as follows:

\[\mathcal{L} = \lambda_1 \mathcal{L}_{3D} + \lambda_2 \mathcal{L}_{2D} + \lambda_3 \mathcal{L}_{v}\]

Here, $\lambda_1$, $\lambda_2$, and $\lambda_3$ are hyperparameters used to balance the contributions of each individual loss function during training.

\section{Experimental Results}

\label{sec:Experimental}
\begin{table*}
  \centering
  \resizebox{0.9\textwidth}{!}{%
  \begin{tabular}{c c c c c c c c}\hline
    \toprule
     \multicolumn{1}{c}{\multirow{2}{*}{}} & \multirow{2}{*}{Method}& \multirow{2}{*}{Output-type} & \multicolumn{3}{c}{3DPW} & \multicolumn{2}{c}{Human3.6M} \\
    \cmidrule(lr){4-6}
    \cmidrule(lr){7-8}
    \multicolumn{1}{c}{}& & & MPVPE$\downarrow$ & MPJPE$\downarrow$ & PA-MPJPE$\downarrow$ & MPJPE$\downarrow$ & PA-MPJPE$\downarrow$\\
    \midrule
    & Graphormer ~\cite{lin2021mesh} & vertices & 87.7 & 74.7 & 45.6 & 51.2 & 34.5 \\
    &METRO ~\cite{lin2021end} & vertices  & 88.2 & 77.1 & 47.9 & 54.0 & 36.7 \\
    &Hybrik-X ~\cite{li2023hybrik}& parameters &94.5  & 80.0 & 48.8 & -- & --  \\
    &Pymaf-X ~\cite{zhang2023pymaf} & parameters  & 110.1 & 92.8 & 58.9 & 57.7 & 40.5 \\
    &Potter ~\cite{zheng2023potter} & parameters & 87.4 & 75.0 & 44.8 & 56.5 & 35.1 \\
    &Fastmetro ~\cite{cho2022cross} & vertices  &84.1 & 73.5 & 44.6 & 52.2 & 33.7 \\
    &PointHMR ~\cite{kim2023sampling} & vertices  & 85.5 & 73.9 & 44.9 & 48.3 & 32.9 \\

    &HMMR ~\cite{kanazawa2019learning} & parameters & 139.3 & 116.5 & 72.6 & --  & 56.9  \\
    &VIBE ~\cite{kocabas2020vibe} & parameters & 99.1 & 82.9 & 51.9 & 65.6 & 41.4 \\
    &TCMR ~\cite{choi2021beyond} & parameters & 111.3 & 95.0 & 55.8 & 62.3 & 41.1 \\
    &MAED ~\cite{wan2021encoder} & parameters & 92.6 & 79.1 & 45.7 &  56.4 & 38.7 \\
    &MPS-Net ~\cite{wei2022capturing} & parameters & 109.6 & 91.6 & 54.0 & 69.4& 47.4 \\
    &FeatER ~\cite{zheng2023feater} & parameters  & 86.9 & 73.4 & 45.9 & \textbf{49.9} & \textbf{32.8} \\
    &GLoT ~\cite{shen2023global} & parameters & 96.3 & 80.7 & 50.6 & 67.0& 46.3 \\
    &\textbf{Ours} & vertices & \textbf{81.9} &
    \textbf{72.7} & \textbf{44.1} & 52.8 & 33.4 \\    
    \bottomrule
  \end{tabular}}
  \caption{We compare our model with the SOTA methods for 3D mesh reconstruction, i.e., Graphormer, METRO, Hybrik-X, Pymaf-X, Potter, PointHMR, HMMR, VIBE, MeshGraphormer, FastMETRO, DeFormer on 3DPW and Human3.6M.}
  \label{tab:t1}

\end{table*}

\subsection{Training}

We conducted all experiments using PyTorch on a system equipped with an Intel(R) Xeon(R) Platinum 8358 CPU (2.60 GHz) and two NVIDIA A100 GPUs. Model parameters were initialized with random values. Training employed the AdamW optimizer for 100 epochs with an initial learning rate of 1e-5, which decayed by a factor of 10 after 10 epochs. Our method processes video sequences as input, utilizing 8 frames per sequence. We batched the training data by grouping 16 sequences together. Prior to training, each sequence was cropped to focus on the human region and resized to a resolution of 224x224 pixels.

\subsection{Datasets and metrics}

We trained our model on two publicly available datasets: Human3.6M~\cite{Human36} and 3DPW~\cite{3DPW}. To improve performance on 3DPW, we further fine-tuned the model using its dedicated test dataset. For Human3.6M evaluation, we employed the standard P2 protocol as described in ~\cite{Human36}. We assessed model performance on both datasets using three established intra-frame metrics: MPVPE, MPJPE, and PA-MPJPE.

\subsection{Comparison of previous approaches}

Our experiments focus on the 3DPW dataset. Table ~\ref{tab:t1} compares the performance of RemoCap with state-of-the-art methods~\cite{shen2023global,zheng2023feater,wei2022capturing,wan2021encoder,choi2021beyond,kocabas2020vibe,kanazawa2019learning,kim2023sampling,cho2022cross,zheng2023potter,zhang2023pymaf,li2023hybrik} on the 3DPW and Human3.6M datasets using MPJPE, PA-MPJPE, and MPVPE metrics. RemoCap outperforms all methods on 3DPW, particularly in MPVPE.

Both Fastmetro~\cite{cho2022cross} and GLoT~\cite{shen2023global} attempt to address occlusion by masking the input data. Fastmetro utilizes masked vertex modeling, while GLoT~\cite{shen2023global} leverages a temporal modeling approach. These methods encourage the learning of inter-frame correlations, which can partially recover occluded poses. However, their reliance solely on intra-frame information limits their effectiveness in handling complex occlusions, particularly those arising during motion restoration. RemoCap tackles this limitation by employing sequence shuffling for diverse input simulations. This strategy exposes the model to a wider range of occlusion scenarios. Furthermore, RemoCap utilizes SD and MD to separate obscured target features from different dimensions within the data. This dissociation facilitates the reconstruction of occluded regions and leads to more accurate pose recovery.

On the 3DPW dataset, RemoCap demonstrates significant improvements over existing methods, particularly when dealing with severe occlusion. Compared to Fastmetro~\cite{cho2022cross},\textbf{ RemoCap achieves reductions of 2.52 mm, 0.82 mm, and 0.54 mm in MPVPE, MPJPE, and PA-MPJPE}, respectively. Similarly, RemoCap outperforms GLoT~\cite{shen2023global} by a substantial margin across all three metrics. These results highlight the effectiveness of RemoCap in tackling the challenging task of 3D human body mesh reconstruction under complex occlusion conditions.

To our surprise, our proposed method didn't exhibit strong performance on the Human3.6M dataset. Upon closer examination, we found that ~\cite{zheng2023feater} preserves the intrinsic structure of feature map representations when incorporating attention, allowing it to perform exceptionally well on the Human3.6M dataset with relatively mild occlusion issues. However, as shown in the appendix figure, our approach still maintains significant competitiveness in the reconstruction process of occluded regions.

\begin{table*}

  \centering
  \resizebox{0.8\textwidth}{!}{%
  \begin{tabular}{c c c c c c c}\hline
    \toprule
    \multirow{2}{*}{Method} & \multicolumn{3}{c}{3DPW} & \multicolumn{2}{c}{Human3.6M} \\
    \cmidrule(lr){2-4}
    \cmidrule(lr){5-6}
    & MPVPE$\downarrow$ & MPJPE$\downarrow$ & PA-MPJPE$\downarrow$ & MPJPE$\downarrow$ & PA-MPJPE$\downarrow$\\
    \midrule
    % \cmidrule(lr){1-1}
    Graphormer ~\cite{lin2021mesh} & 87.7 & 74.7 & 45.6 & 51.2 & 34.5 \\
    % \cmidrule(lr){1-7}
    METRO ~\cite{lin2021end}       & 88.2 & 77.1 & 47.9 & 54.0 & 36.7 \\
    % \cmidrule(lr){1-7}
    PointHMR ~\cite{kim2023sampling} & 85.5 & 73.9 & 44.9 & 48.3 & 32.9 \\
    % \cmidrule(lr){1-7}
    Fastmetro ~\cite{cho2022cross} &84.1 & 73.5 & 44.6 & 54.0 & 34.0  \\
    % \cmidrule(lr){1-7}
    \textbf{Ours} & \textbf{81.9} & \textbf{72.7} & \textbf{44.1} & \textbf{52.8} & \textbf{33.4} \\
    \bottomrule
  \end{tabular}}
    \caption{We compared RemoCap to the leading model-free method using the Video-base testing protocol. This evaluation demonstrates RemoCap's superior intra-frame accuracy, particularly in terms of MPVPE.}
  \label{tab:t2}
  \vspace{-4mm}
\end{table*}

\textbf{Model-free approach:}Table ~\ref{tab:t2} compares RemoCap to the model-free method Fastmetro~\cite{cho2022cross}. RemoCap significantly outperforms Fastmetro~\cite{cho2022cross} on the challenging 3DPW dataset with complex backgrounds, \textbf{achieving improvements of 2.2 mm, 0.8365 mm, and 0.5 mm in MPVPE, MPJPE, and PA-MPJPE}, respectively. Even on the Human3.6M dataset with simpler backgrounds, \textbf{RemoCap maintains strong performance, demonstrating enhancements in MPJPE (1.2370 mm) and PA-MPJPE (0.5 mm)}.These improvements stem from RemoCap's ability to decouple spatial information of target features and capture existing motion information. This allows RemoCap to effectively reconstruct occluded body parts, leading to superior performance across various background complexities.

\begin{table*}
  \centering
  \resizebox{0.9\textwidth}{!}{%
  \begin{tabular}{c|c c c c c c}\hline
    \toprule
     \multicolumn{1}{c}{\multirow{2}{*}{}} & \multirow{2}{*}{Method} & \multicolumn{3}{c}{3DPW} & \multicolumn{2}{c}{Human3.6M} \\
    \cmidrule(lr){3-5}
    \cmidrule(lr){6-7}
    \multicolumn{1}{c}{}& & MPVPE$\downarrow$ & MPJPE$\downarrow$ & PA-MPJPE$\downarrow$ & MPJPE$\downarrow$ & PA-MPJPE$\downarrow$ \\
    \midrule
    \multirow{5}{*}{Video-based}& VIBE ~\cite{kocabas2020vibe} & -- & 91.9 & 57.6  & 78.0 & 53.3  \\
    &MEVA ~\cite{luo20203d}& -- & 86.9 & 54.7 & 76.0 & 53.2 \\
    &TCMR ~\cite{choi2021beyond}& 102.9 & 86.5  & 52.7 & 73.6 & 52.0 \\
    &MPS-Net ~\cite{wei2022capturing} & 99.7  & 84.3 & 52.1 & 69.4 & 47.4 \\
    &GLoT ~\cite{shen2023global} & 96.3 & 80.7 & 50.6 & 67.0 & 46.3  \\
    \cmidrule(lr){1-7}
    \multirow{2}{*}{Video-based Model-Free}&\textbf{Our(w/o post-pro)} & \ \textbf{81.9} & \ \textbf{72.7} & \ \textbf{44.1} & \ \textbf{52.8} & \ \textbf{33.4}  \\
    \multirow{1}{*} &\textbf{Our(w post-pro) ~\cite{zeng2022smoothnet}} & -- & 74.5 & 45.9 & 64.0 & 45.2 \\
    \bottomrule
  \end{tabular}}
    \caption{We compare our processed model with video-based approaches using a comprehensive set of metrics, including MPVPE, MPJPE and PA-MPJPE. This approach is consistent with prior studies, such as ~\cite{shen2023global}.}
  \label{tab:t3}
  \vspace{-6mm}
\end{table*}

\textbf{Video-based approach:}Table ~\ref{tab:t3} compares RemoCap to other video-based methods. RemoCap achieves superior performance across all metrics. While methods like GLoT ~\cite{shen2023global} and MPSNet~\cite{wei2022capturing} attempt to model temporal information, they focus either on long-range or short-range dependencies. This limits their ability to effectively utilize motion cues within the video sequence to address occlusions. In contrast, RemoCap excels at capturing the continuity of motion features for the target object. By decoupling local motion information, RemoCap achieves significant improvements in all three metrics (MPVPE, MPJPE, PA-MPJPE) on both datasets. The complex backgrounds of the 3DPW dataset present a greater challenge, but RemoCap still demonstrates substantial improvements \textbf{(14.4 mm, 8.0 mm, and 6.5 mm)}. Even on the simpler backgrounds of Human3.6M, RemoCap \textbf{maintains strong performance with enhancements of 14.2 mm and 12.5 mm in MPJPE and PA-MPJPE}, respectively. By effectively capturing the target object's motion and addressing feature coupling limitations, RemoCap surpasses previous methods in 3D human body mesh reconstruction.

% \begin{figure}[hb]
%   \centering
%     \includegraphics[width=0.4\linewidth]{final_qua.png} 
% \caption{The image shows the result of our model effectively handling occlusion and complex actions, Comparison with advanced parametric modeling methods ~\cite{zhang2023pymaf}. }
% \label{fig:t6}
% \end{figure}

\begin{figure*}[h]
  \centering
    \includegraphics[width=0.85\linewidth]{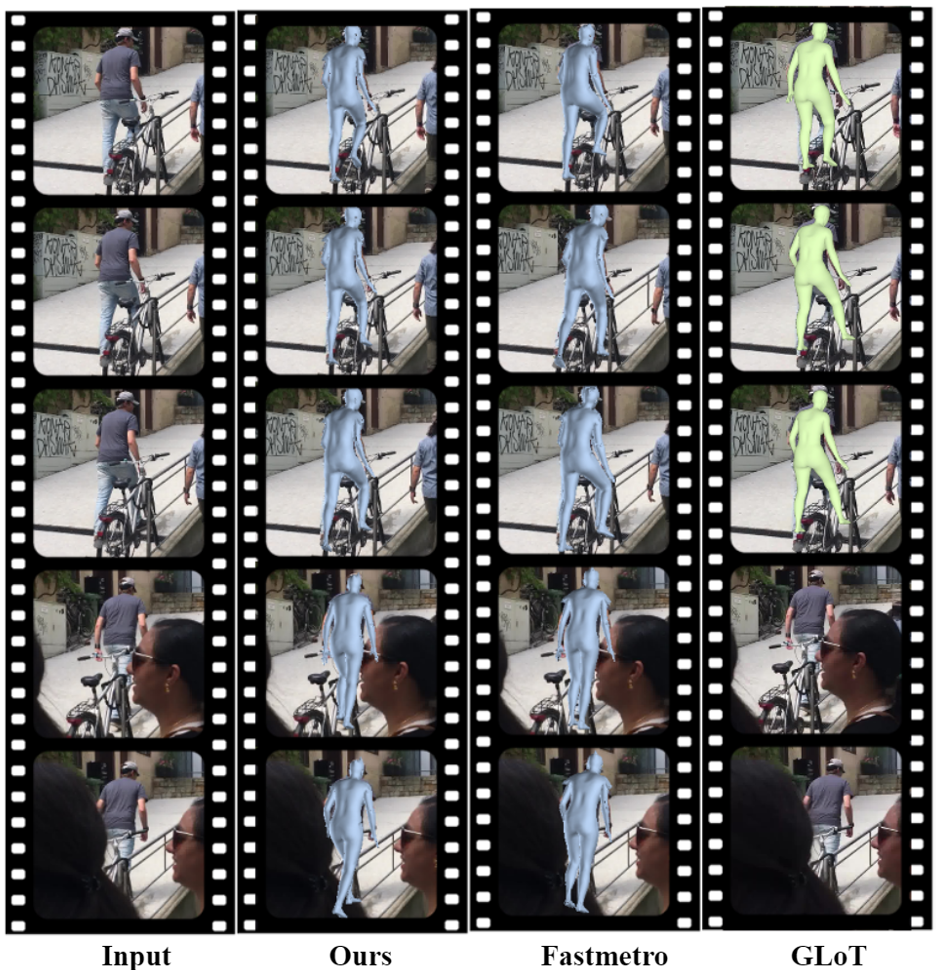} 
\caption{The image presents a comparative analysis of the performance of different algorithms, showcasing their ability to handle complex scenes such as a person riding a bicycle. The original video sequence is shown on the far left, serving as a benchmark. Progressing to the right, the processed outputs from three different algorithms are displayed: Our method, Fastmetro ~\cite{cho2022cross}, and GLoT ~\cite{shen2023global}.}
\label{fig:t5}
\end{figure*}

Figure ~\ref{fig:t5} compares the motion reconstruction performance of RemoCap with Fastmetro~\cite{cho2022cross} and GLoT~\cite{shen2023global} during occlusion in the leg region. Notably, both ~\cite{cho2022cross} and ~\cite{shen2023global} exhibit significant reconstruction errors at frames 5 and 3, respectively.

\begin{figure*}[h]
  \centering
    \includegraphics[width=0.85\linewidth]{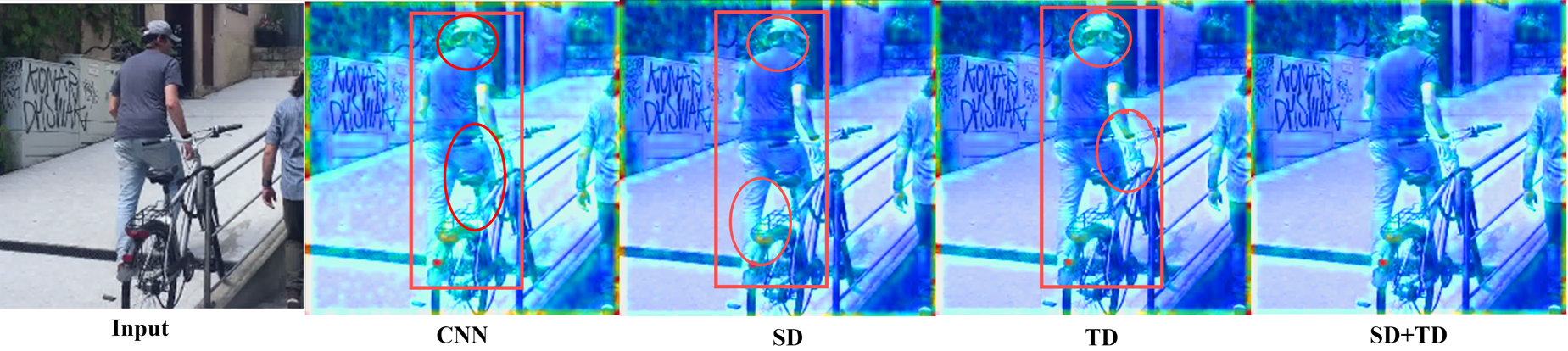} 
\caption{We visually compare the features extracted by the original CNN with the heatmaps before and after using the SD and the TD modules.}
\label{fig:t4}
\vspace{-3mm}
\end{figure*}

\subsection{Qualitative evaluation}

While existing methods offer advantages, they have limitations when dealing with occlusion.  Fastmetro~\cite{cho2022cross} prioritizes fast convergence by reducing interactions between input tokens, but this can neglect the impact of motion-induced occlusion. Conversely, GLoT~\cite{shen2023global} focuses on capturing motion relationships to address global offsets, but it struggles with occluded objects. RemoCap overcomes these limitations by considering both motion and spatial information, allowing for effective reconstruction even in challenging occlusion scenarios.

RemoCap tackles occlusion challenges through its SD and MD modules. As illustrated in Figure ~\ref{fig:t5}, \textbf{the SD module isolates features of the occluded object (\textit{e.g.}, male subject's legs) from the background (bike).} Simultaneously, \textbf{the MD module focuses on capturing motion information within the target object.} This decoupling process allows RemoCap to reconstruct occluded motion parts with stability, even in complex scenarios. By simultaneously considering both motion information and the target object's spatial features, RemoCap effectively addresses occlusion during reconstruction. \textbf{This dual focus enables RemoCap to surpass existing methods in terms of robustness during the reconstruction process.}

In summary, RemoCap demonstrates significant improvements in 3D human body mesh reconstruction, achieving superior performance across MPVPE, MPJPE, and PA-MPJPE metrics. Notably, RemoCap excels in handling challenging occluded scenarios, showcasing its robustness and generalizability. These achievements not only highlight RemoCap's potential for various reconstruction tasks but also establish it as a powerful and efficient new paradigm for model-free reconstruction methods.

\subsection{Ablation experiment}

Table ~\ref{tab:t4} quantifies the contributions of RemoCap's SD and MD modules in addressing feature coupling. On the challenging 3DPW dataset with complex backgrounds, the SD module's decoupling of target features leads to a significant improvement of 0.45 $mm$ in PA-MPJPE. The MD module, designed to capture motion information, contributes to a substantial increase of 1.97 $mm$ in MPVPE. The combined effect of these modules allows RemoCap to achieve optimal performance on 3DPW. Even on the Human3.6M dataset with simpler backgrounds and missing vertex labels, RemoCap demonstrates progress in PA-MPJPE. This highlights the effectiveness of SD and MD in tackling feature coupling and achieving SOTA performance on various reconstruction tasks.

\begin{table}[h]
  \centering
  
  \scalebox{0.7}{
  \begin{tabular}{c c| c c c c c}\hline
    \toprule
       \multicolumn{1}{c}{\multirow{2}{*}{SD}} &  \multicolumn{1}{c}{\multirow{2}{*}{TD}} & \multicolumn{3}{c}{3DPW} & \multicolumn{2}{c}{Human3.6M} \\  
     \cmidrule(lr){3-5}
     \cmidrule(lr){6-7}
      & \multicolumn{1}{c}{} & MPVE$\downarrow$ & MPJPE$\downarrow$ &PA-MPJPE$\downarrow$ & MPJPE$\downarrow$ & PA-MPJPE$\downarrow$ \\
    \midrule
    $\times$ & $\times$ & 84.1 & 73.5 & 44.6 & \bf{52.2} & 33.7\\
    $\surd$ & $\times$ & 82.5 & 73.1 & 44.2 & 52.9 & 33.5\\
    $\times$ & $\surd$ & 82.1 & 73.1 & 44.5 & 52.9 & 34.6\\
    \textbf{$\surd$} & \textbf{$\surd$} & \ \bf{81.9} & \ \bf{72.7 } & \ \bf{44.1} & 52.8 & \ \bf{33.4} \\
    \bottomrule
   \end{tabular}}
  \caption{Investigates the effectiveness of SD module and TD module in RemoCap.}
  \label{tab:t4}
  \vspace{-4mm}
\end{table}

Figure ~\ref{fig:t4} visualizes heatmap features within the reconstruction window, showcasing the effects of different methods (CNN, SD, and MD). Compared to CNN, the SD module \textbf{significantly reduces non-target features in the body region.} This effectively separates the spatial features of the target object from the combined feature space. MD builds upon SD's success. \textbf{It suppresses features from static body parts} like the head and neck, which exhibit minimal motion. Conversely, MD preserves features in highly mobile regions like the hand. Additionally, \textbf{MD effectively reduces motion-related features from non-target objects around the hand}. This demonstrates MD's ability to isolate motion information within the target object's feature space, leading to a more refined representation.

\section{Conclusion}
\label{sec:Conclusion}

This paper introduces RemoCap, a groundbreaking framework for robust and accurate human mesh vertex recovery. RemoCap addresses a critical challenge in this field: occlusion. It achieves this through two novel modules, \textbf{S}patial \textbf{D}isentang-
lement (\textbf{SD}) and \textbf{M}otion \textbf{D}isentanglement (\textbf{MD}), which significantly improve performance in occluded scenarios. Additionally, RemoCap tackles the issue of temporal speed loss, ensuring consistent results across video sequences. Experimental results demonstrate RemoCap's effectiveness in reducing 3D human mesh distortion and vertex jitter, especially in occluded scenarios. Furthermore, RemoCap achieves significant improvements in human pose estimation accuracy. Notably, RemoCap surpasses the state-of-the-art on the challenging 3DPW benchmark dataset.We believe RemoCap's potential extends beyond the tested scenarios. Its robust design suggests that it will deliver similar outstanding performance in various real-world occlusion situations.

\newpage
\bibliographystyle{ieeenat_fullname}
\bibliography{main}

\clearpage
\newpage

\appendix
\vspace*{1em}{\centering\Large\bf%
Appendix
\vspace*{1.5em}}

\begin{figure*}[htbp]
    \centering
    \includegraphics[width=0.85\textwidth]{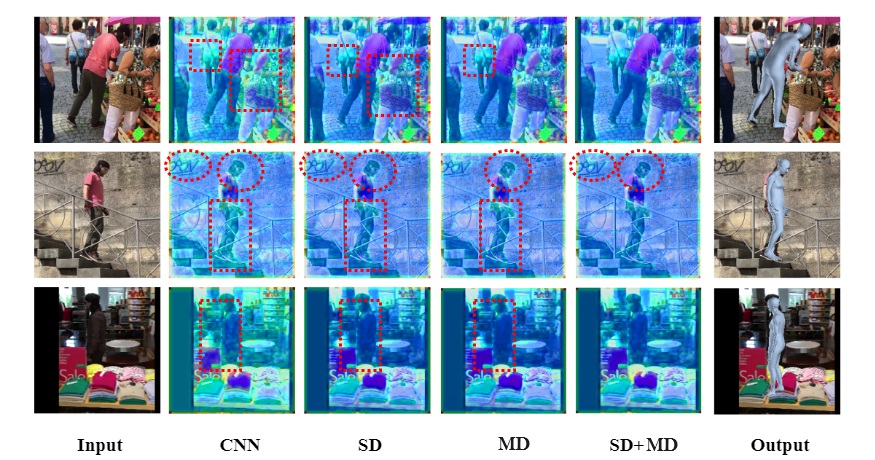}
    \caption{We visually compare the features extracted by the original CNN with the heatmaps before and after using the SD and MD modules.}
    \label{fig:s10}
\end{figure*}

\begin{figure*}[htbp]
    \centering
    \includegraphics[width=0.85\textwidth]{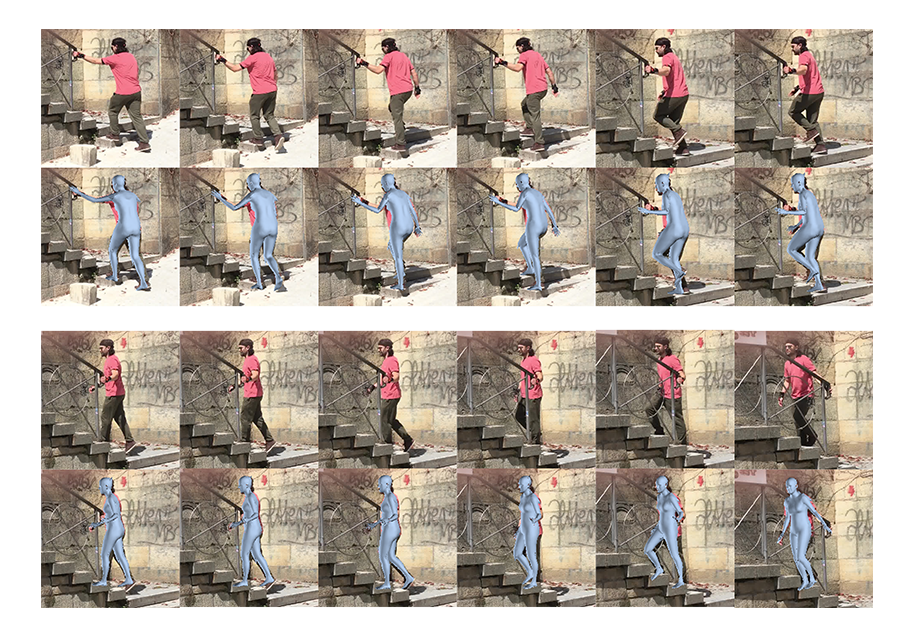}
    \caption{The figure shows the performance of our model in a sequential video task, showing the stability of the reconstruction results, the stability against occlusion interference, and the advantages of limb position alignment. For a more detailed video demonstration click on our project page: \url{https://wanghongsheng01.github.io/RemoCap/}.}
    \label{fig:s9}
\end{figure*}

\section{Datasets}
This section introduces two key datasets employed for 3D human pose estimation: 3DPW~\cite{3DPW} and Human3.6M~\cite{Human36}.

\textbf{3DPW: }A comprehensive collection of multi-person 3D poses captured in real-world scenarios with high variability. Unlike controlled settings, 3DPW pushes the boundaries by including diverse, uncontrolled environments. The dataset features 60 videos recorded at 30 fps using a phone, making it closer to real-world applications.  IMU sensors are utilized to obtain near ground-truth SMPL parameters~\cite{loper2015smpl} (pose and shape) for multiple individuals in each frame.  Activities include walking, sitting, and interactions with objects.  Additionally, synchronized multi-view images and 2D pose annotations are provided.

\textbf{Human3.6M: }This established benchmark dataset captures diverse human activities in a controlled indoor environment using high-resolution video. Human3.6M offers a valuable resource for researchers due to its realism and variety of poses and interactions, including walking, posing, and other daily actions.  The dataset features multiple subjects performing these activities, providing a comprehensive representation of human movements.  Synchronized multi-view videos, 3D joint annotations, and ground-truth camera parameters are included, making it suitable for developing and evaluating 3D human pose estimation algorithms.

\section{Metries}
This section details the evaluation metrics employed to assess the performance of 3D human pose estimation algorithms.

\begin{itemize}
    \item \textbf{MPJPE:} This metric calculates the average Euclidean distance between the predicted and ground-truth 3D joint locations. Lower MPJPE values indicate higher accuracy in pose estimation.
    \item \textbf{PA-MPJPE: } PA-MPJPE refines MPJPE by applying Procrustes Analysis (PA) for 3D alignment. This alignment mitigates the influence of global rotation and translation differences, leading to a more accurate measure of pose similarity.
    \item \textbf{MPVPE:} Similar to MPJPE, MPVPE measures the average Euclidean distance between corresponding vertices on the predicted and ground-truth 3D mesh. It provides a more comprehensive evaluation of mesh reconstruction accuracy.
\end{itemize}

The unit of intra-frame metric is millimeters, and the unit of inter-frame metric is $mm/s^2$.

\section{Feature coupling issue}
This section discusses a critical challenge in 3D human body mesh reconstruction from videos: spatiotemporal feature coupling. This phenomenon refers to the undesirable mixing of features from non-target objects with those of the target person (Figure \ref{fig:s10}). It manifests in two primary ways:

\begin{itemize}
    \item \textbf{Spatial Feature Disentanglement:}  Within a single frame, features from occlusions, background clutter, or other individuals can become entangled with the target's features in the feature space.  During reconstruction, this coupling can lead to the inadvertent incorporation of non-target features, resulting in vertex distortions in the target mesh.
    \item \textbf{Motion Feature Disentanglement:}  The challenge persists across frames. Features disentangled from previous frames accumulate over time.  Additionally, static features of the target (\textit{e.g.}, torso) tend to dominate the feature space across consecutive video segments, overshadowing inter-frame motion features (\textit{e.g.}, limbs).  This dominance hinders the model's ability to extract continuous motion information, leading to: (i)Loss of inter-frame motion details in the reconstructed mesh sequence. (ii)Cumulative reconstruction errors cause jitter and continuous distortions.
\end{itemize}
Effectively disentangling and utilizing valuable feature information from this coupled state is crucial for accurate 3D human body mesh reconstruction.  The key lies in developing methods that can: (i)Eliminate the influence of non-target features in both intra-frame and inter-frame contexts. (ii)Correctly associate target features with the target object.

\section{Result}
\subsection{Stability of reconstruction results}
\begin{figure*}[htbp]
    \centering
    \includegraphics[width=0.85\textwidth]{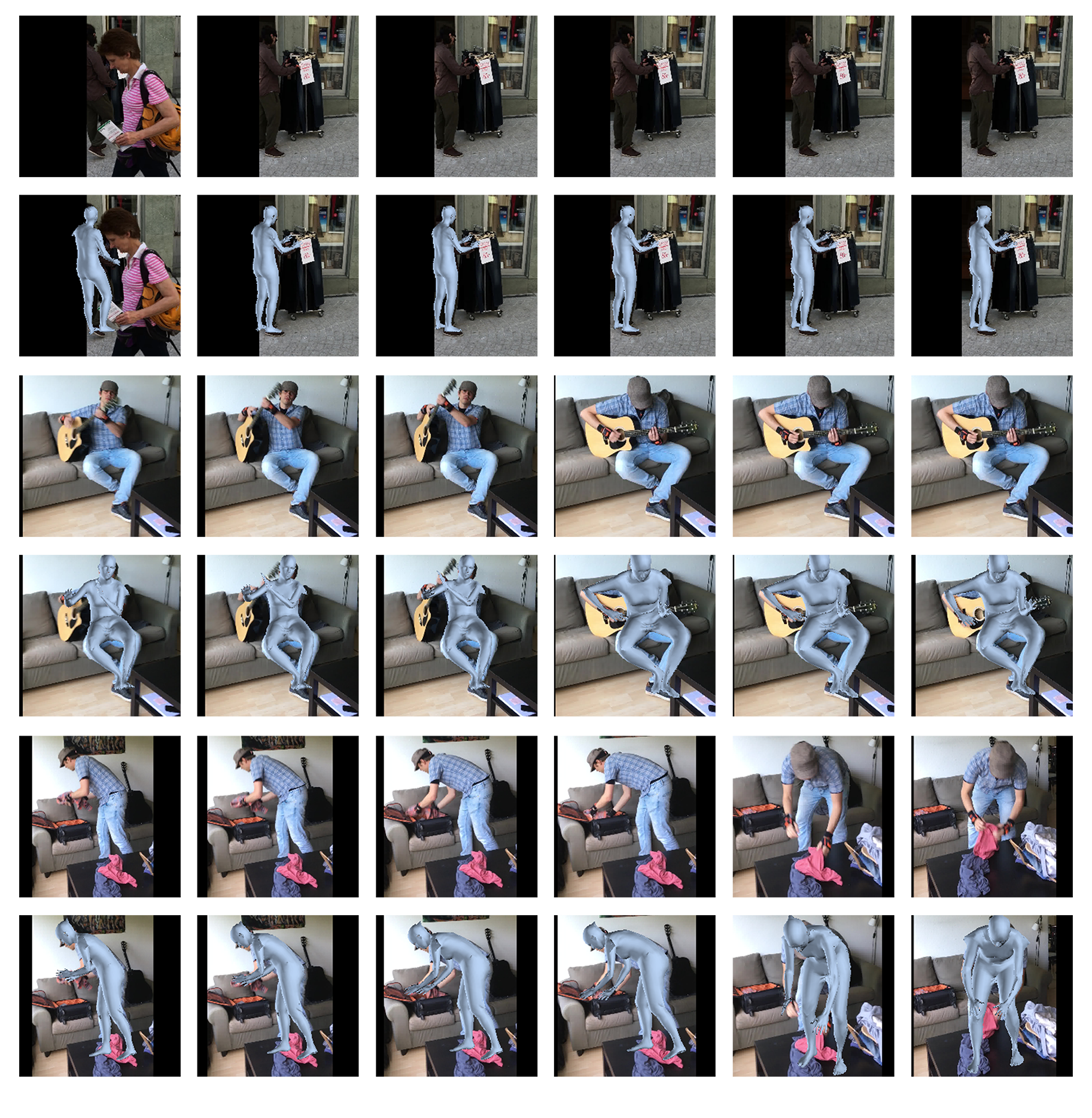}
    \caption{Here we show the recovery results of our method for several sequences. From left to right, they are arranged in chronological order of the frames, which shows the robustness and accuracy of the reconstruction of our method.}
    \label{fig:s15}
\end{figure*}

Figure \ref{fig:s9} showcases the accuracy and stability of our reconstruction method. Even in complex scenarios with occlusions, our approach effectively overcomes these challenges, delivering high-fidelity reconstructions of the target character. Figure \ref{fig:s15} further demonstrates the stability and accuracy of our reconstruction method in handling complex occluded scenes.

\begin{figure*}[htbp]

  \renewcommand\twocolumn[1][]{#1}%
    \centering
    \includegraphics[width=1\textwidth]{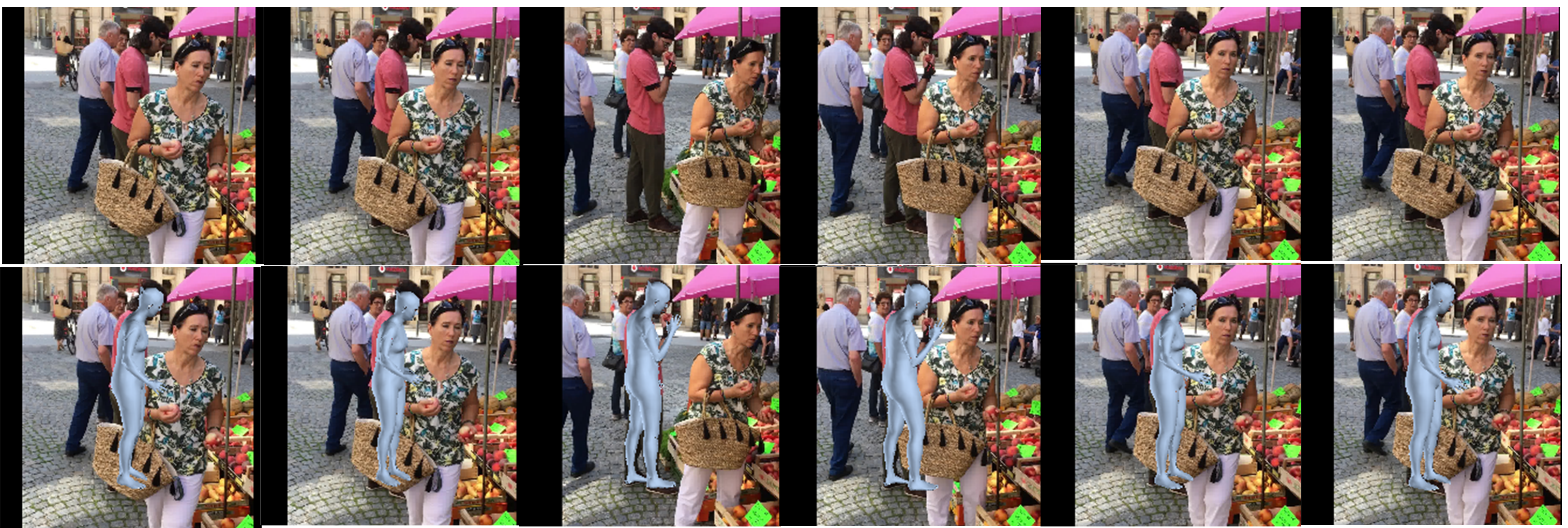}
    \caption{Here we show the excellent reconstruction results of our method in a scene with a mixture of multi-motion targets.}
    \label{fig:s16}
\end{figure*}

\subsection{Reconstruction under occlusion}
\begin{figure}[h]
    \centering
    \includegraphics[width=0.5\textwidth]{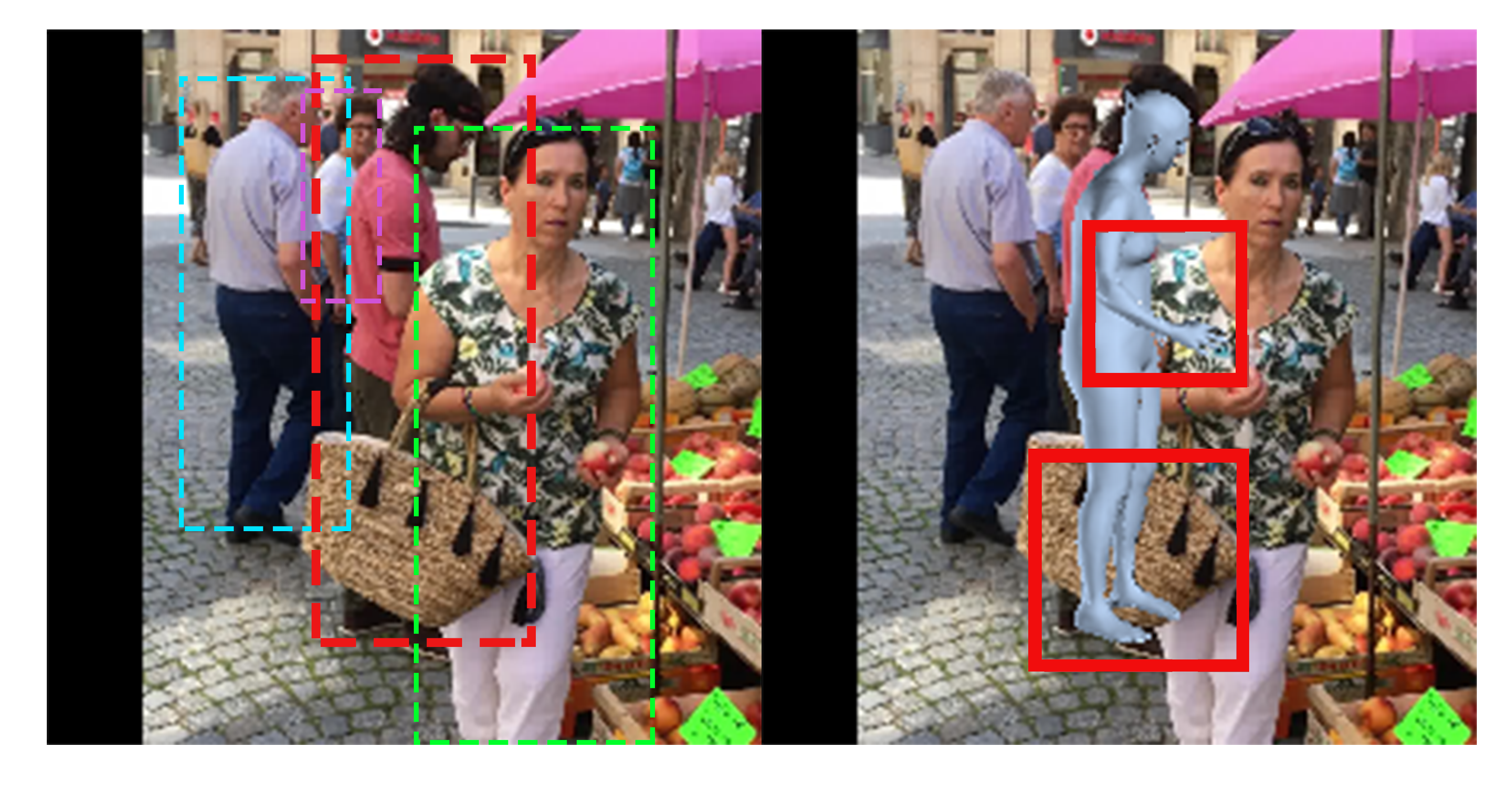}
    \caption{Our method is able to reconstruct the target person accurately in complex scenes dealing with multiple people, without interference from non-target persons. The red box in the left image indicates the main character, and our method can stably reconstruct the occluded part in the right image.}
    \label{fig:s12}
\end{figure}

Our method excels at reconstructing the target person (red box, Figure \ref{fig:s12}) in intricate scenes with multiple people. By highlighting non-target individuals with other colored boxes, Figure \ref{fig:s12} visually demonstrates our method's ability to handle occlusions and focus on the target. Further reinforcing this capability, Figure \ref{fig:s16} showcases RemoCap's full video stream human body recovery. As the figure reveals, our method maintains exceptional performance in reconstructing the target subject even under occlusion.

\subsection{Comparison with state-of-the-art methods}
\begin{figure}[h]
 \renewcommand\twocolumn[1][]{#1}%
    \centering
    \includegraphics[width=1.0\linewidth]{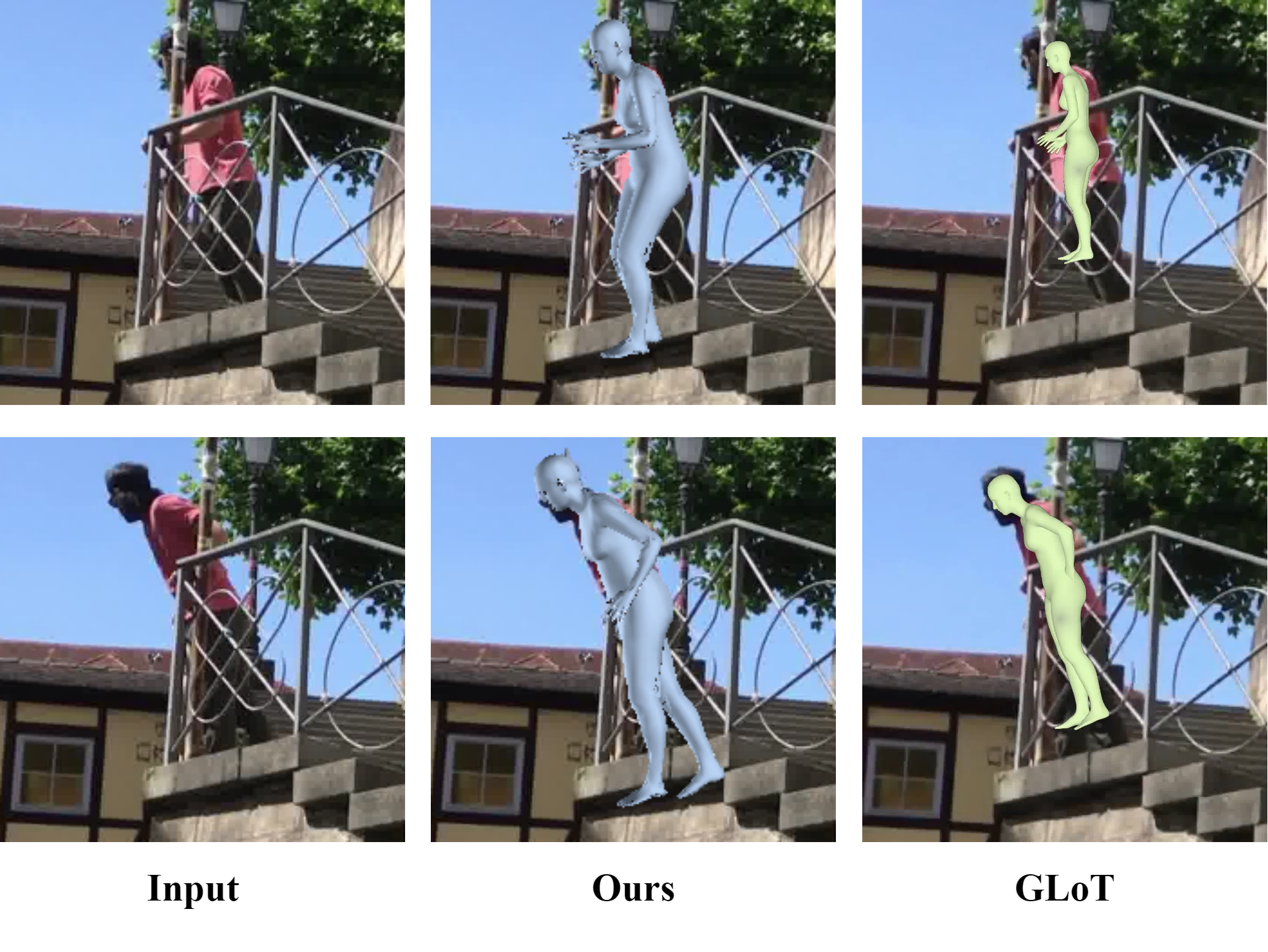}
    \caption{Compared to state-of-the-art video-based methods, we have more obvious advantages in facing complex environments such as occlusion}
    \label{fig:s13}
\end{figure}

Figure \ref{fig:s13} compares the performance of our method against leading algorithms~\cite{shen2023global} (SOTA) in video pose estimation, particularly for complex scenes like cycling (far left, benchmark video). Our method, employing a model-free approach, consistently outperforms the SOTA method (displayed to the right) which relies on parametric models.
This advantage stems from the inherent benefits of our model-free approach and the Transformer architecture itself. Transformers excel at capturing the relationships between individual frames and across frames (local-whole relationship). This stability allows them to effectively handle complex scenarios like occlusions, intrusions from non-target objects, and more.

\section{Ablation experiment}
To visualize the impact of different feature decoupling methods on reconstruction accuracy, Figure \ref{fig:s10} presents the intermediate heatmap features of reconstruction windows learned using three approaches: CNN (baseline), Spatial Detachment (SD), and Motion Detachment (MD).
\begin{itemize}
    \item \textbf{Spatial Detachment (SD): }Compared to the CNN baseline, SD significantly reduces activations for non-target features within the target object's body area. This effectively separates the target's spatial features from the background clutter in the feature space.
    \item \textbf{Motion Detachment (MD):} In contrast to SD,MD focuses on suppressing features from stationary body parts. This is evident in the minimal response in the head and neck regions of the heatmap. However, MD successfully preserves motion features, particularly in areas like the hands. Additionally, MD effectively reduces activations for non-target features around the hands. This demonstrates its ability to isolate motion features within the target's feature space, further enhancing the capture of dynamic movements.
\end{itemize}

\section{Future works}
Given our method's model-free nature, we believe it holds promise for transferring the task to hand reconstruction. This exploration is a promising avenue for future research.
However, a key limitation of our current approach is its high computational cost due to the model-free design. The reported results in this paper employed two A100 GPUs for training. To address this, we plan to investigate incorporating temporal information strategically. The goal is to reduce the model's computational burden by allowing it to process only a relevant subset of data within the sequence.
			        
\end{CJK}
\end{document}